\documentclass[11pt, a4paper, logo, copyright, nonumbering]{kwaipilot}
\usepackage{dblfloatfix}
\usepackage{ulem}
\usepackage{caption}
\usepackage{dramatist}
\usepackage{xspace}
\usepackage{pifont} %
\usepackage{multirow}
\usepackage{tcolorbox}
\usepackage{xltabular}
\usepackage{longtable}
\usepackage{hyperref}
\usepackage{booktabs}
\usepackage{array}
\usepackage{xcolor}
\definecolor{lightgray}{RGB}{110,110,110}
\usepackage{hyperref}
\usepackage{footnote}
\usepackage{geometry}
\usepackage{tcolorbox}
\usepackage{caption}
\usepackage{rotating}
\usepackage{graphicx}
\usepackage{subcaption}
\usepackage{float}
\usepackage{amsfonts}
\usepackage{amsmath}
\usepackage{amssymb}
\usepackage{lineno}
\usepackage{multirow}
\usepackage{adjustbox}

\usepackage[bottom]{footmisc}

\usepackage{CJKutf8}
\usepackage{setspace}

\usepackage{dsfont}
\usepackage{array} %
\usepackage{tabularx} %
\usepackage{xcolor} %
\usepackage{tabularx}
\usepackage{booktabs}

\usepackage{lipsum}  %
\usepackage{multicol} %

\usepackage{indentfirst} 
\usepackage[sort&compress,numbers]{natbib}

\usepackage{makecell}

\usepackage{titlesec}
\titlespacing*{\paragraph}{0pt}{0.3ex plus 0.2ex minus .2ex}{1em}

\makeatletter
\def\@BTrule[#1]{%
  \ifx\longtable\undefined
    \let\@BTswitch\@BTnormal
  \else\ifx\hline\LT@hline
    \nobreak
    \let\@BTswitch\@BLTrule
  \else
     \let\@BTswitch\@BTnormal
  \fi\fi
  \global\@thisrulewidth=#1\relax
  \ifnum\@thisruleclass=\tw@\vskip\@aboverulesep\else
  \ifnum\@lastruleclass=\z@\vskip\@aboverulesep\else
  \ifnum\@lastruleclass=\@ne\vskip\doublerulesep\fi\fi\fi
  \@BTswitch}
\makeatother

\addto\extrasenglish{
}

 {\begin{list}{}%
         {\setlength{\leftmargin}{#1}}%
         \item[]%
 }
 {\end{list}}
 
\reportnumber{001} %

\title{\centering KAT-V1: Kwai-AutoThink Technical Report }

\author{
    Zizheng Zhan$^{*}$$^{\dag}$, Ken Deng$^{*}$, Huaixi Tang$^{*}$, Wen Xiang$^{*}$, Kun Wu$^{*}$, Weihao Li, Wenqiang Zhu, Jingxuan Xu, Lecheng Huang, Zongxian Feng, Shaojie Wang, Shangpeng Yan, Xuxing Chen, Jiaheng Liu, Zhongyuan Peng, Zuchen Gao, Haoyang Huang, Xiaojiang Zhang, Jinghui Wang, Zheng Lin, Mengtong Li, Huiming Wang, Ziqi Zhan, Yanan Wu, Yuanxing Zhang, Jian Yang, Guang Chen, Haotian Zhang, Bin Chen, Bing Yu \\
    Kwaipilot Team \\
    \texttt{\{zhanzizheng, dengken, zhanghaotian\}@kuaishou.com} \\
}

\begingroup
\footnotetext{
$^*$Equal contribution. $^\dag$Corresponding author. }
\endgroup

\renewcommand{\arraystretch}{1.2} 

\begin{abstract}
We present Kwaipilot-AutoThink (KAT), an open-source 40B large language model developed to address the overthinking problem in reasoning-intensive tasks,
where an automatic thinking training paradigm is proposed to dynamically switch between reasoning and non-reasoning modes based on task complexity.
Specifically, first, we construct the dual-regime dataset based on a novel tagging pipeline and a multi-agent synthesis strategy,
and then we apply Multi-Token Prediction (MTP)-enhanced knowledge distillation, enabling efficient and fine-grained reasoning transfer with minimal pretraining cost. 
Besides, we implement a cold-start initialization strategy that introduces mode-selection priors using majority-vote signals and intent-aware prompting. Finally, we propose Step-SRPO, a reinforcement learning algorithm that incorporates intermediate supervision into the GRPO framework, offering structured guidance over both reasoning-mode selection and response accuracy. Extensive experiments across multiple benchmarks demonstrate that KAT consistently matches or even outperforms current state-of-the-art models, including DeepSeek-R1-0528 and Qwen3-235B-A22B, across a wide range of reasoning-intensive tasks while reducing token usage. 
Notably, KAT outperforms all open-source models and even surpasses o3-mini on the leakage-controlled LiveCodeBench Pro.
Beyond academic evaluation, KAT has been successfully deployed in Kwaipilot (i.e., Kuaishou's internal coding assistant), where it improves real-world development workflows with high accuracy, efficiency, and controllable reasoning behaviors. Moreover, we are actively training a 200B Mixture-of-Experts (MoE) model with 40B active parameters, and early results already show significant gains, further demonstrating the scalability of the AutoThink paradigm.
\end{abstract}

\begin{document}

\maketitle

\begin{figure}[htbp]
    \centering
    \includegraphics[width=1\linewidth]{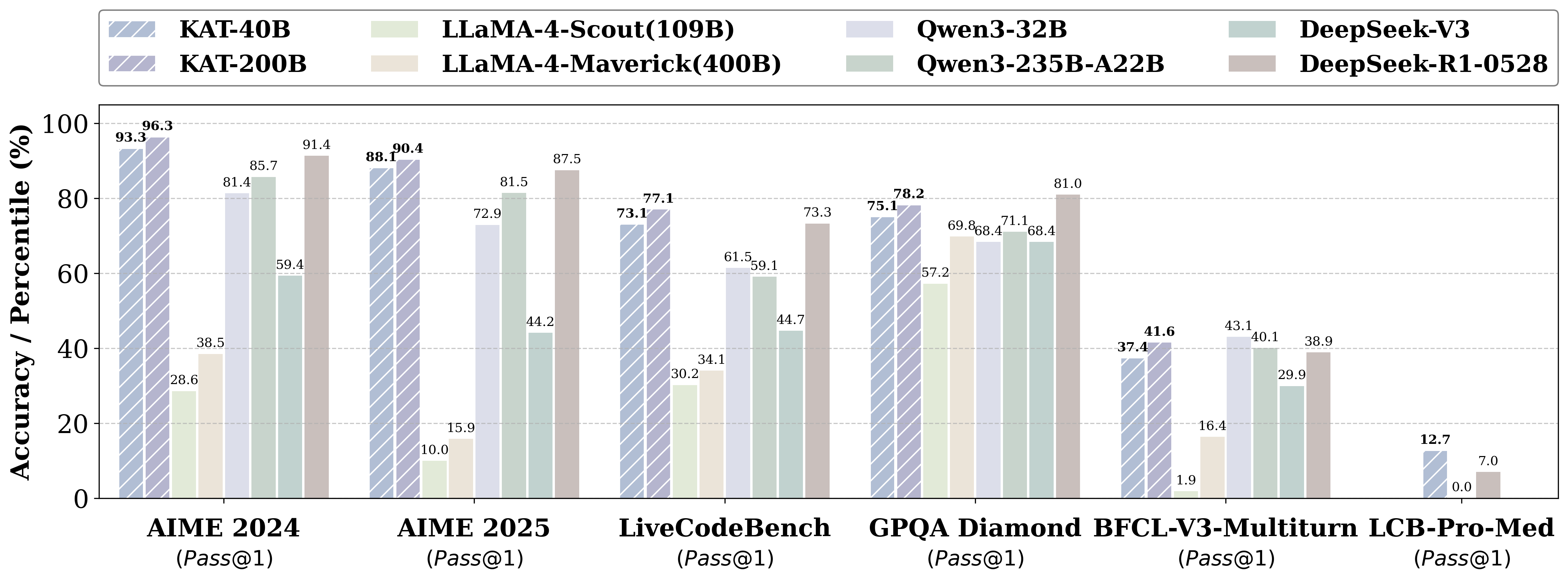}
    \caption{Performance of Kwaipilot-AutoThink on various benchmarks.}
    \label{fig:ScoresOfModels}
\end{figure}

\newpage
\section{Introduction}
Large language models (LLMs) have recently achieved impressive, even human-level, performance across a broad spectrum of natural language understanding and generation tasks \cite{guo2025deepseekR1, liu2024deepseek-v2, bi2024deepseek-llm, guo2024deepseek-coder, yang2025qwen3, qwen2025qwen2.5, yang2024qwen2, bai2023qwen1, meta2025llama4, grattafiori2024llama3, touvron2023llama2, wang2024mtu-Bench, liu2025comprehensiveSurvey, liu2024m2rc-eval, deng2024r2c2-coder}. A key driver of this progress is the ability of LLMs to perform chain-of-thought (CoT) reasoning, which has substantially improved their problem-solving abilities on complex tasks.

However, recent studies have revealed that applying CoT reasoning indiscriminately to all inputs--regardless of their inherent complexity--often results in significant computational overhead, increased latency, and excessive token consumption, particularly for straightforward queries \cite{chen2024do-Not-Think-That-Much, fatemi2025concise-reasoning}. This phenomenon, commonly referred to as overthinking, has emerged as a critical bottleneck for the real-world deployment of LLMs in interactive applications, where both efficiency and responsiveness are essential\cite{sui2025stopOverthinking, lou2025adacot}.

To mitigate the overthinking problem, a growing body of research has focused on improving the efficiency by adaptively reasoning in large language models \cite{jiang2025thinkOnlyWhenYouNeed, fang2025thinkless, lou2025adacot, tu2025learningWhenToThink, zhang2025adaptthink}. One of the representative approaches is to empower models with the ability to dynamically decide whether to engage in reasoning, based on the complexity of the input. Notable examples include AdaptThink \cite{zhang2025adaptthink} and AdaCoT \cite{lou2025adacot}, which employ reinforcement learning \cite{shao2024deepseekmath, schulman2017proximal-Poplicy-Optimization-Algorithms} to train models that can selectively trigger CoT reasoning during inference. Despite their promising results, most of these efforts still face two fundamental limitations. First, they typically lack sufficient supervision to help models develop robust, query-specific preferences over reasoning strategies. Second, existing RL frameworks are often coarse-grained, providing limited intermediate feedback and failing to model the step-wise nature of reasoning and mode selection.

Moreover, research on knowledge distillation \cite{gou2021knowledge-Distillation-a-survey, hinton2015distilling-The-Knowledge-in-a-neural-network, liu2024ddk} has aimed to improve model efficiency by transferring reasoning capabilities from stronger teacher models to smaller or less expensive student models. However, conventional distillation approaches often require extensive pretraining or rely solely on final-answer alignment, which fails to capture the nuanced utility of intermediate reasoning steps. Although methods like Multi-Token Prediction (MTP) \cite{liu2024deepseek-V3-technical-report} have been proposed to model future utility, few have explored how MTP can be systematically integrated into the distillation pipeline for reasoning tasks.

Building upon these insights,
as shown in Figure~\ref{fig:Introduction},
we present \textbf{Kwaipilot-AutoThink (KAT)}, a 40B open-source LLM that addresses the overthinking challenge through a two-stage training framework (i.e., pre-training and post-training stages) and achieves adaptive reasoning, efficient knowledge injection, and real-world applicability as follows:

\begin{figure}
    \centering
    \includegraphics[width=1\linewidth]{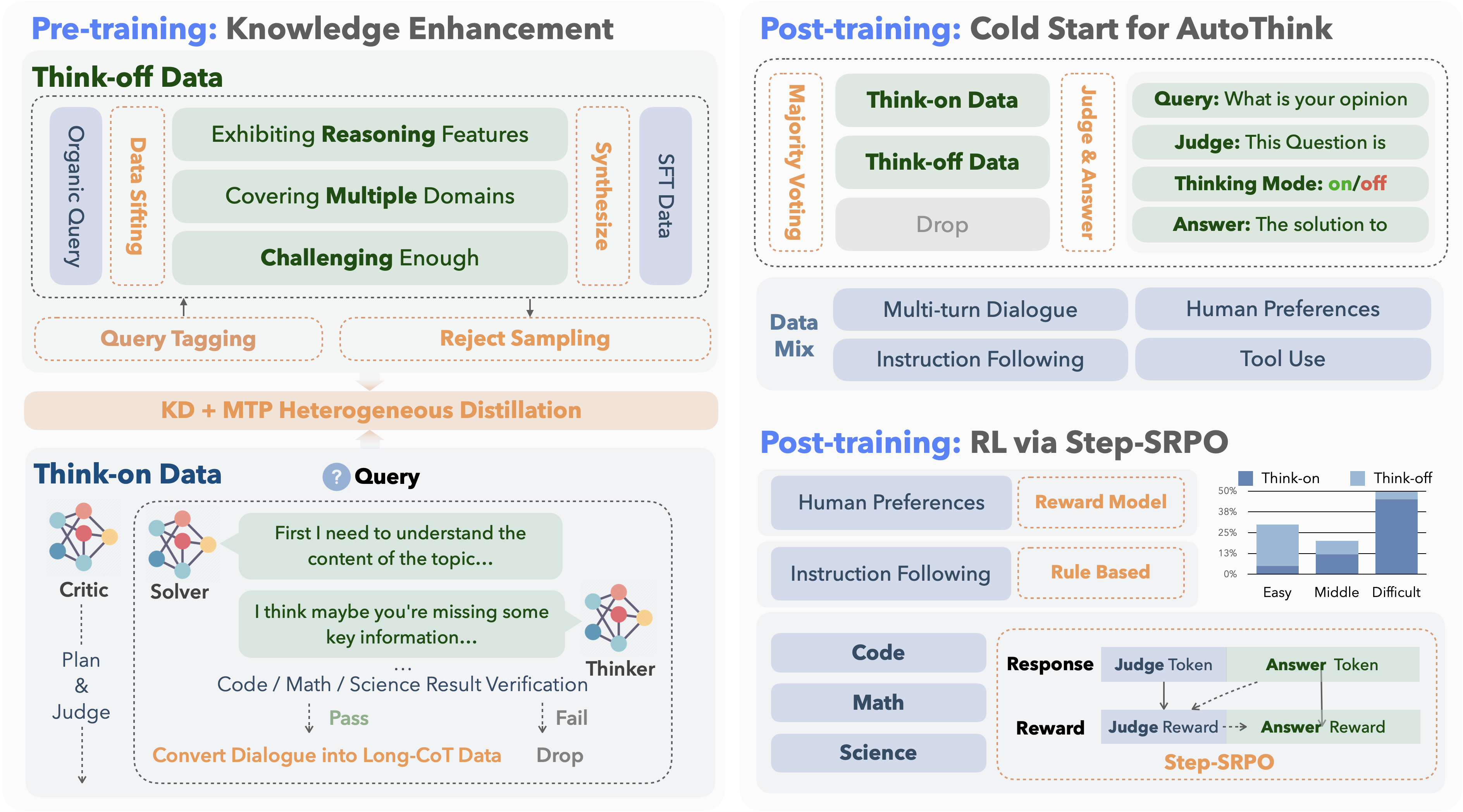}
    \caption{The \textbf{two-stage} AutoThink training framework for controllable and efficient reasoning in large language models. }
    \label{fig:Introduction}
\end{figure}

In the pre-training stage\textsuperscript{1}, we introduce a knowledge enhancement strategy based on knowledge distillation and multi-token prediction. Specifically, we propose a novel data labeling system to curate a diverse and challenging set of training queries. Leveraging this system, we construct two distinct data regimes--Think-off data synthesized via our tagging system, and Think-on data generated through a multi-agent framework grounded in state-of-the-art LLMs. Unlike prior work, we combine knowledge distillation with Multi-Token Prediction (MTP) to train on this dual-regime data, enabling more granular utility modeling and efficient acquisition of high-quality reasoning capabilities while avoiding the cost of extensive pretraining.

\footnotetext[1]{In this paper, pre-training refers to continued pre-training based on the upscaled Qwen2.5-32B to a 40B scale.}

In the post-training stage, we introduce an auto-thinking method to achieve efficient token usage for reasoning LLMs. Specifically, we first establish a robust cold-start pipeline for AutoThink training. For each query, we employ majority voting to determine the appropriate thinking mode and condition generation on both query attributes and inferred user intent. 
Then, we propose Step-SRPO, an enhanced reinforcement learning algorithm that introduces intermediate supervision into the SRPO \cite{zhang2025SRPO} framework, enabling structured guidance at two key levels: (1) accurate thinking-mode selection, and (2) final response correctness under the selected mode. This staged design better aligns the model's internal decision process with the complexity of the input, stabilizing training and improving convergence.

Through this two-stage paradigm, Kwaipilot-AutoThink not only achieves state-of-the-art performance across multiple benchmarks-including math, code, commonsense, and scientific tasks---but also significantly reduces token consumption. For instance, our model achieves scores of 88.1 and 93.3 on AIME2025 and AIME2024, respectively, while using fewer tokens than the current state-of-the-art LLM. 
Notably, Kwaipilot-AutoThink ranks first among all open-source models on LiveCodeBench Pro\cite{liveCodeBench-Pro}, a challenging benchmark explicitly designed to prevent data leakage, and even surpasses strong proprietary systems such as Seed\cite{Seed1.6_Bytedance} and o3-mini\cite{OpenAI_o3mini_2025}.
Furthermore, Kwaipilot-AutoThink demonstrates robust performance in real-world deployment: it has been integrated into Kwaipilot, Kuaishou's internal code assistant, where it effectively handles a wide range of development tasks, underscoring the real-world practicality of our thinking control framework and cost-efficient training approach.

In summary, our contributions are fourfold:

\begin{itemize}
    \item \textbf{Comprehensive Data Synthesis with MTP-Enhanced Knowledge Distillation:} We develop a versatile data synthesis framework that integrates both non-reasoning and long-chain reasoning regimes. This is combined with a Multi-Token Prediction (MTP)-enhanced knowledge distillation strategy, enabling fine-grained and efficient transfer of reasoning capabilities without requiring costly full-scale pretraining.
    \item \textbf{AutoThink with Intermediate Supervision:} We introduce AutoThink with Intermediate Supervision, a staged reinforcement learning framework (Step-SRPO) that imposes structured guidance on both reasoning mode selection and final response generation. This approach enhances convergence and strengthens the model's ability to align its reasoning behavior with task complexity.
    \item \textbf{State-of-the-Art Performance with Efficiency Gains:} Kwaipilot-AutoThink achieves state-of-the-art results across multiple benchmarks. Remarkably, despite having only 40 billion parameters, our model matches or surpasses the performance of leading LLMs with several times more parameters. For example, Kwaipilot-AutoThink scores leads all open-source models and even outperforms proprietary competitors such as Seed and o3-mini on LiveCodeBench Pro. These results demonstrate that our model delivers both competitive performance and high efficiency, underscoring the effectiveness of our training paradigm, reasoning control framework, and the fine-grained data mixture strategy that balances domain diversity and data complexity.
    \item \textbf{Real-World Deployment and Practicality:} Kwaipilot-AutoThink has been successfully deployed in Kwaipilot, Kuaishou's internal coding assistant. It shows robust performance across diverse software development scenarios, validating the practical value of our adaptive thinking mechanism and cost-effective training strategy in real-world environments.
\end{itemize}

Together, these contributions advance the frontier of controllable and efficient reasoning in large language models, laying a solid foundation for scalable, deployable, and high-performing AI systems in both academic and industrial settings.

\section{Architecture: Upscaling from 32B to 40B}

Building on recent studies that explore the relationship between model depth and performance, we scaled the Qwen2.5-32B model to a 40B-parameter variant via a targeted layer-wise expansion strategy. Specifically, we conducted a layer saturation analysis---measuring the cosine similarity between input and output token representations at each transformer layer, following methodologies introduced in~\cite{young2024yi-OpenFoundationModel}---to identify redundant or saturated layers (i.e., layers where representational change is minimal). This diagnostic revealed that approximately one-fourth of the model's layers exhibited near-identity transformations, suggesting a limited contribution to feature abstraction.

To enhance model capacity while preserving training efficiency, we selectively upscaled these saturated layers by duplication and continued training with diverse, high-quality data spanning both reasoning-intensive and general tasks. In parallel, infrastructure optimizations revealed that the 40B parameter scale achieves peak Model FLOPs Utilization (MFU) of up to 50\%, providing an optimal balance between scale and computational efficiency. The resulting model, \textbf{KAT-V1-40B}, demonstrates state-of-the-art zero-shot and few-shot performance among models below the 40B parameter threshold. Detailed model configurations can be found in Table~\ref{tab:autothink-arch}.

\begin{table}[ht]
  \centering
  \caption{Model architecture details for Kwaipilot--AutoThink.}
  \label{tab:autothink-arch}
  \begin{tabular}{lcccc}
    \toprule
    \textbf{Model} & \textbf{Layers} & \textbf{Tie Embedding} & \textbf{Heads (Q / KV)} & \textbf{Context Length} \\
    \midrule
    Kwaipilot--AutoThink--40B & 80 & No & 40 / 8 & 32K \\
    \bottomrule
  \end{tabular}
\end{table}

\section{Pre-training: Knowledge Enhancement}

This stage marks the first step toward realizing the AutoThink paradigm, where the model is trained to follow externally provided think-mode instructions, enabling it to alternate between reasoning and non-reasoning behaviors based on input prompts. Specifically, we adopt trigger tokens such as \texttt{<think\_on>} and \texttt{<think\_off>} during training to indicate whether reasoning should be activated for each query. These tokens are not involved in loss computation, but serve as effective mode selectors during training.

This mechanism is aligned with prior works such as Llama-Nemotron~\cite{bercovich2025llama-nemotron} and Qwen3~\cite{yang2025qwen3}, which also rely on external cues to control reasoning behavior. Although the model does not yet autonomously determine when to reason, this approach already addresses the overthinking problem observed in fully CoT-activated models by enabling passive switching between reasoning modes. It thus lays the foundation for more advanced capability development in subsequent stages, where dynamic mode selection becomes increasingly model-driven.

\subsection{Knowledge Distillation with Multi-Token Prediction}

\begin{figure}[H]
    \centering
    \includegraphics[width=1\linewidth]{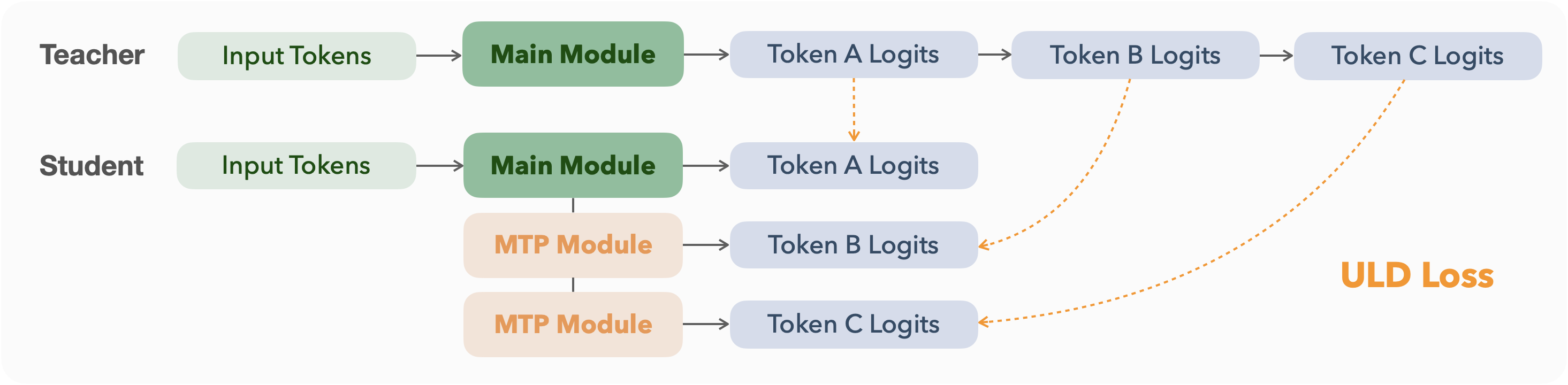}
    \caption{Illustration of our heterogeneous knowledge distillation framework combining Multi-Token Prediction (MTP) and Universal Logit Distillation Loss (ULD Loss).}
    \label{fig:knowledge-distillation}
\end{figure}

Technically, we proposed a heterogeneous distillation framework that combines Multi-Token Prediction (MTP) with a Universal Logit Distillation Loss (ULD Loss) to enhance the effectiveness of knowledge transfer from a larger teacher model to a smaller student model. As illustrated in Figure~\ref{fig:knowledge-distillation}, both the teacher and student models process the same input tokens, while the student model incorporates additional MTP modules to predict future tokens (e.g., Token~B, Token~C) in parallel with the main decoding pathway.
To accommodate the different reasoning requirements of our dual-regime training data, we employed DeepSeek-V3 and DeepSeek-R1-0528 as teacher models for Think-off (non-reasoning) and Think-on (reasoning) data, respectively. These models were chosen due to their strong open-source performance and extensive community validation, making them well-suited for transferring high-quality knowledge in their respective domains.

Among various intermediate alignment strategies (e.g., embedding-level, MLP activations, LM head outputs), we empirically found that aligning token-level logits yielded the most effective transfer performance. The ULD Loss enables supervision to flow from the teacher's sequential logits (Token~A, B, C) to the student's heterogeneous MTP modules, even when their prediction architectures differ. This allows for flexible and architecture-agnostic distillation.
By predicting multiple future tokens, the MTP modules implicitly model ``future rewards'' and broaden the supervision space, while ULD Loss bridges the gap between sequential and parallel predictions. This design significantly improves the efficiency of knowledge injection, and enables rapid cold-start initialization with minimal computational overhead.

\subsection{Data Construction and Distribution}
To effectively enhance the capabilities of KAT-V1-40B, we curate a dual-regime dataset that is logically rich, cross-domain, and sufficiently challenging. This process begins with a novel data labeling system that leverages state-of-the-art LLMs to assess both the difficulty and the domain characteristics of each query. Queries are then classified into Think-off and Think-on categories, based on their intrinsic complexity and the availability of verifiable answers.
For Think-off queries, we leverage DeepSeek-V3 to generate responses that do not require multi-step reasoning. Meanwhile, we employ multiple state-of-the-art LLMs to perform reject sampling, thereby ensuring high-quality outputs by selecting only the most accurate and relevant completions. In contrast, Think-on queries are processed through a multi-agent framework, composed of a solver, a thinker, and a critic. The solver provides initial answers, the thinker reflects on and iteratively improves the solution, and the critic supervises the full process to ensure logical consistency and output quality. Only verified outputs from this pipeline are retained and converted into long-CoT data.

\begin{figure}
    \centering
    \includegraphics[width=1\linewidth]{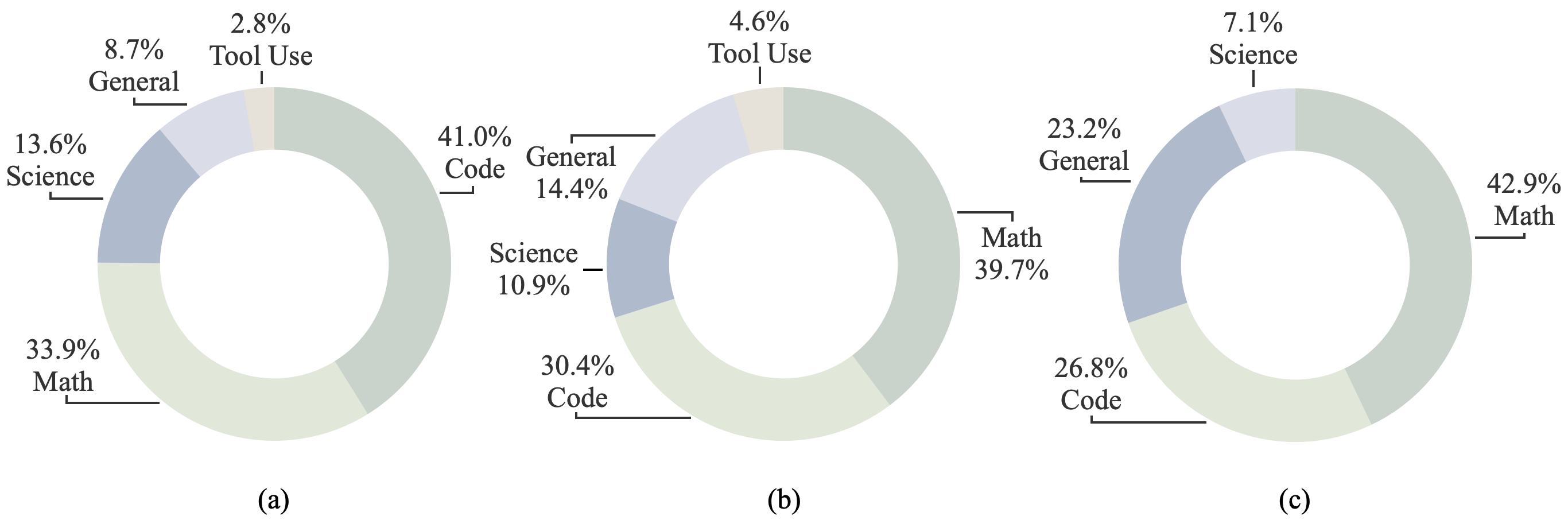}
    \caption{Domain distributions of training data used in Stage 1 (a), Stage 2 (b), and Stage 3 (c) of the KAT training pipeline.}
    \label{fig:dataConstrution}
\end{figure}

Our Stage 1 training corpus comprises approximately 10 million examples, carefully curated to balance reasoning complexity and response efficiency. Among these, 34.8\% of the data is annotated as Think-on, while 65.2\% is labeled as Think-off, establishing a practical equilibrium between complex reasoning and lightweight generation. As illustrated in Figure~\ref{fig:dataConstrution}, the dataset covers a diverse range of domains. This broad domain coverage enables the model to generalize its reasoning capabilities across a wide variety of real-world tasks and instruction types.

\subsection{Analysis: Effectiveness of MTP-Enhanced Distillation}

To evaluate the impact of our Multi-Token Prediction (MTP)-enhanced knowledge distillation, we compare it against standard knowledge distillation (KD) across several benchmarks. Our results show that integrating MTP with KD leads to approximately a 5 percent improvement in performance across these benchmarks.

\section{Post-training: Auto-Think}
\subsection{Cold Start}
Building upon the large-scale knowledge enhancement achieved through the upscaled model and the efficient training strategy, we now take a critical step forward: empowering the model to autonomously determine the appropriate reasoning mode based on the input query. While the previous stage enabled passive mode switching via explicit tags, it still relied on externally provided instructions.

To overcome this limitation, we design a cold-start initialization mechanism that serves as the bridge from passive compliance to active decision-making. By introducing intent-aware analysis, majority-vote distillation, and carefully filtered supervision signals, we instill in the model an initial inductive bias to recognize whether a query warrants deep reasoning or a direct response. This transition marks a foundational advance in the AutoThink paradigm---shifting the model's behavior from externally guided reasoning to internally inferred reasoning mode selection, setting the stage for reinforcement-based fine-tuning in the next phase.

\subsubsection{Data Construction}

As illustrated in Figure~\ref{fig:Introduction}, we construct our training data through a structured pipeline designed to explore and supervise the model's preference between Think-on and Think-off modes.
Given a user query, we employ a majority voting mechanism across multiple model outputs to determine the preferred reasoning mode---either Think-on or Think-off. For the selected reasoning path, we sample corresponding responses using two distinct models: DeepSeek-R1 for the Think-On mode and DeepSeek-V3 for the Think-Off mode. This ensures a diverse and task-aligned generation process.

To enhance the model's understanding of reasoning mode selection, we introduce an auxiliary evaluation step. Specifically, for each query-response pair, we use DeepSeek-V3 to generate a judgment rationale explaining the appropriateness of the selected mode. This step provides the training signal necessary for learning the alignment between query complexity and reasoning depth.
Furthermore, to ensure coverage and avoid overfitting to biased query distributions, 1\% of the queries are randomly assigned a mode (regardless of majority voting), enforcing the model's exposure to diverse reasoning scenarios.

The final training samples are formatted in a unified structure that encapsulates both the judgment and the answer generation process. We define two templates as shown in Table~\ref{tab:template} and Table~\ref{tab:special_tokens}.

This format enables the model to jointly learn when to engage in deeper reasoning and how to perform reasoning-aware answer generation. The explicit separation of reasoning analysis and final answer encourages structured alignment between mode selection, justification, and response quality.

\begin{table}[htbp]
\centering
\caption{
\textbf{Formatting templates used in the final training data.} The Think-On and Think-Off modes follow a unified structure comprising judgment and answer segment.
}
\small
\begin{tabular}{>{\raggedright\arraybackslash}p{0.25\textwidth}|>
{\raggedright\arraybackslash}p{0.25\textwidth}}
\toprule
\textbf{Think-On Mode} & \textbf{Think-Off Mode} \\
\hline
\texttt{<judge>} \newline
\textcolor{blue}{\{judge\_analysis\}} \newline
\texttt{</judge>} \newline
\newline
\texttt{<think\_on>} \newline
\texttt{<think>} \newline
\textcolor{red}{\{thinking\_content\}} \newline
\texttt{</think>} \newline
\newline
\texttt{<answer>} \newline
\textcolor{blue}{\{response\}} \newline
\texttt{</answer>}
&
\texttt{<judge>} \newline
\textcolor{blue}{\{judge\_analysis\}} \newline
\texttt{</judge>} \newline
\newline
\texttt{<think\_off>} \newline
\texttt{<answer>} \newline
\textcolor{blue}{\{response\}} \newline
\texttt{</answer>}
\\
\bottomrule
\end{tabular}
\label{tab:template}
\end{table}

\begin{table}[htbp]
\centering
\caption{\textbf{Special tokens and their descriptions.}}
\label{tab:special_tokens}
\small
\begin{tabular}{ll} 
\toprule
\textbf{Special Token}           & \textbf{Description} \\
\midrule
\texttt{<judge>}        & Analyzes input query to determine whether reasoning is required. \\
\texttt{<think\_on/off>} & Specifies whether reasoning should be activated ("on") or skipped ("off"). \\
\texttt{<think>}         & Marks the beginning of reasoning in "think-on" mode. \\
\texttt{<answer>}        & Marks the beginning of the model's answer. \\
\bottomrule
\end{tabular}
\end{table}

\subsubsection{Data Distribution}

The composition of our dataset spans six distinct categories: \textbf{code, math, science, general, multi-turn, and tool use}. As illustrated in Figure~\ref{fig:dataConstrution}, this distribution ensures broad coverage across both reasoning-intensive and general-purpose tasks, enabling the model to learn robust mode selection strategies under varying task complexities.

\paragraph{Code, Math, and Science}
We constructed a challenging question corpus from diverse public and proprietary sources, applying rigorous selection criteria to ensure broad coverage, disciplinary diversity, and balanced difficulty levels. Samples with verifiable results or test cases were prioritized. For mathematical and scientific questions meeting inclusion criteria but lacking definitive answers, we annotated soft labels using robust open-source models via majority voting. Difficulty levels were annotated granularly, and the dataset distribution was calibrated through stratified sampling. Excessively difficult questions were oversampled, with multiple responses collected per instance. Finally, response validity underwent multi-step verification, including rule-based validation, model-based response comparison, and sandboxed execution.

\paragraph{General and Multi-turn Dialogue}
We collected a diverse corpus of instructions spanning a broad spectrum of topics. Each instruction was annotated with multi-dimensional labels and difficulty ratings, followed by clustering and filtering. Particular attention was paid to balancing instances involving reasoning mode transitions within multi-turn dialogues. To preserve the model's inherent preferences for reasoning patterns, we employed a prompt-based approach leveraging intermediate checkpoints to collect model responses via rejection sampling. For each query, responses underwent human preference evaluation using a reward model across multiple dimensions, retaining the optimal response.

\paragraph{Tool Use}
To enhance the model's capability in agent-based scenarios, we specifically collected data related to tool use, task planning, and reflective reasoning. We gathered diverse tool sets and multi-turn interaction trajectories from both public and internal sources, covering a wide range of scenarios. Targeting real-world application settings, we further collect interaction trajectories in software development contexts via multi-agent systems. Furthermore, we optimized the model's tool calling scheme for end-to-end agent operation, which requires it to perform thorough reasoning and provide explanations before executing any calls.

In total, the Stage 2 training corpus comprises approximately 3.5 million examples, spanning the aforementioned categories, with the proportion of Think-on to Think-off examples approximately 2:1. As illustrated in Figure~\ref{fig:dataConstrution}, the distribution maintains a balanced coverage across domains such as code, math, general reasoning, multi-turn dialogue, and tool-use trajectories, ensuring that the model is exposed to both structured reasoning and open-ended interaction patterns.

\subsection{Step-SRPO}

Typically, post-training of large language models (LLMs) involves separate reinforcement learning (RL) stages targeting different, often mutually exclusive, objectives. This sequential strategy may lead models to oscillate between conflicting objectives, causing instability similar to a ``seesaw'' effect. To address this challenge, we propose the \textbf{Step-SRPO} framework (the detailed AutoThink RL implementation will be released in a forthcoming companion paper), which unifies multiple reinforcement learning objectives into a single, coherent training session. Specifically, Step-SRPO aims to simultaneously achieve three interrelated goals.

\paragraph{Thinking Control.} 
Although KAT acquires initial automatic reasoning abilities from the AutoThink cold-start training (Stage 2), these capabilities remain unstable, influenced by variables such as query domain and length. Therefore, subsequent reinforcement learning is necessary to stabilize and refine the model's capability to autonomously activate explicit reasoning (``think-on'') for complex queries and deactivate it (``think-off'') for simpler tasks.

\paragraph{Reasoning RL.}
While we achieved strong initial performance during the cold-start phase, standalone reasoning-based RL training, particularly with challenging query-verifier pairs, effectively improves the model's reasoning capability. However, this singular approach tends to bias the model towards producing answers predominantly in the think-on mode to maximize reward signals, thus negatively impacting its AutoThink gating ability. Incorporating reasoning-focused RL into the Step-SRPO structure mitigates this bias, ensuring balanced reasoning activation based on query complexity.

\paragraph{General RL.}
General RL aims to broadly enhance the model's robustness and reliability across diverse scenarios through reward signals covering multiple tasks. Yet, applying general RL alone may degrade KAT's AutoThink discriminative performance, necessitating integration within the Step-SRPO framework to maintain a balanced AutoThink capability.

\subsubsection{Algorithm}
Unlike conventional language models that apply uniform reasoning depth across all queries, Step SRPO introduces a "reasoning necessity assessment" phase. This allows models to determine whether deep reasoning is required before generating responses, achieving a dynamic balance between computational efficiency and output quality.
The Step-SRPO reward structure comprises two interrelated components:

\paragraph{Judge Reward} Calculated based on the overall correctness of the model's reasoning activation decisions, reinforcing accurate gate predictions across the entire query-response cycle.

\paragraph{Answer Reward} Primarily determined by the correctness and quality of the final answer itself, further modulated by the Judge Reward. This ensures that the answer quality remains consistently aligned with the effectiveness of the reasoning decision.

This structured reward mechanism directs the model toward contextually appropriate reasoning utilization, balancing computational efficiency with response accuracy.

\subsubsection{Reward}

\paragraph{Math} Responses are evaluated by symbolic equivalence checks (algebraic simplification and numeric approximation). Correct answers receive full reward; incorrect or absent solutions score zero.

\paragraph{Code} Responses labeled as code are evaluated by running predefined unit tests. Passing all tests yields maximal reward; any test failure yields zero.

\paragraph{Science} For multiple-choice questions, we extract the first letter of the submitted option and compare against the reference key. A match scores one; a mismatch scores zero.

\paragraph{General} Use lightweight heuristics (keyword overlap, simple classifier) to assign a binary correct/incorrect signal.

\subsubsection{Data Distribution}

Maintaining an appropriate distribution of data difficulty is critical to the effectiveness of Step-SRPO in reinforcing the model's AutoThink capability. If the query-verifier pairs are too simple or too difficult, the model may fall into undesirable local optima---such as consistently avoiding reasoning or over-engaging in unnecessary reasoning. Conversely, a dataset consisting solely of moderately difficult instances may lead to unstable reasoning-mode preferences, preventing the model from learning robust decision boundaries.

To ensure robust learning dynamics, we constructed a dataset with broadly distributed difficulty levels, reflecting a spectrum from low to high complexity, while biasing toward harder examples that require nuanced reasoning. The difficulty distribution of this 45K-sample training corpus is illustrated in Figure~\ref{fig:Introduction}, which shows a deliberate concentration in the high-difficulty region while retaining a non-negligible proportion of easier examples for stabilization.
In addition to difficulty, domain coverage also plays a pivotal role in shaping AutoThink behavior. During reasoning-focused reinforcement learning, data is often skewed toward math and code, which risks overfitting the model to domain-specific reasoning shortcuts or ``hacks.'' To mitigate this, we curated a dataset spanning math, code, and general-purpose reasoning queries. Each query-verifier pair was assigned a difficulty score based on the model's current capability, and the final dataset maintains a broad difficulty distribution with a bias toward higher-difficulty samples. A small portion of lower-difficulty examples was also included to support general reinforcement learning objectives and stabilize training dynamics.
As shown in Figure~\ref{fig:dataConstrution}, this dataset maintains domain diversity, preventing over-specialization and encouraging generalizable AutoThink behavior.

\subsection{Analysis}

To rigorously evaluate the effectiveness of AutoThink reasoning control, we conduct a comprehensive analysis of both training and inference behaviors. Specifically, we focus on two key dimensions: reasoning mode activation (think-on vs. think-off) and token efficiency across training steps and diverse benchmark tasks.

During training, we examine how the model's decision-making evolves under the Step-SRPO framework, tracking the frequency of reasoning-mode activations and the associated output length. These metrics provide insight into how the model learns to balance reasoning depth with efficiency. During inference, we further investigate how reasoning control generalizes across domains, highlighting how reasoning-intensive benchmarks sustain high think-on usage, while others benefit from adaptive reasoning suppression. Collectively, these trends reveal how KAT progressively develops fine-grained, context-sensitive reasoning behavior, delivering strong performance with reduced computational overhead.

\subsubsection{Think-On vs. Think-Off and Token Count Dynamics During Training}

\begin{figure}[H]
    \centering
    \includegraphics[width=1\linewidth]{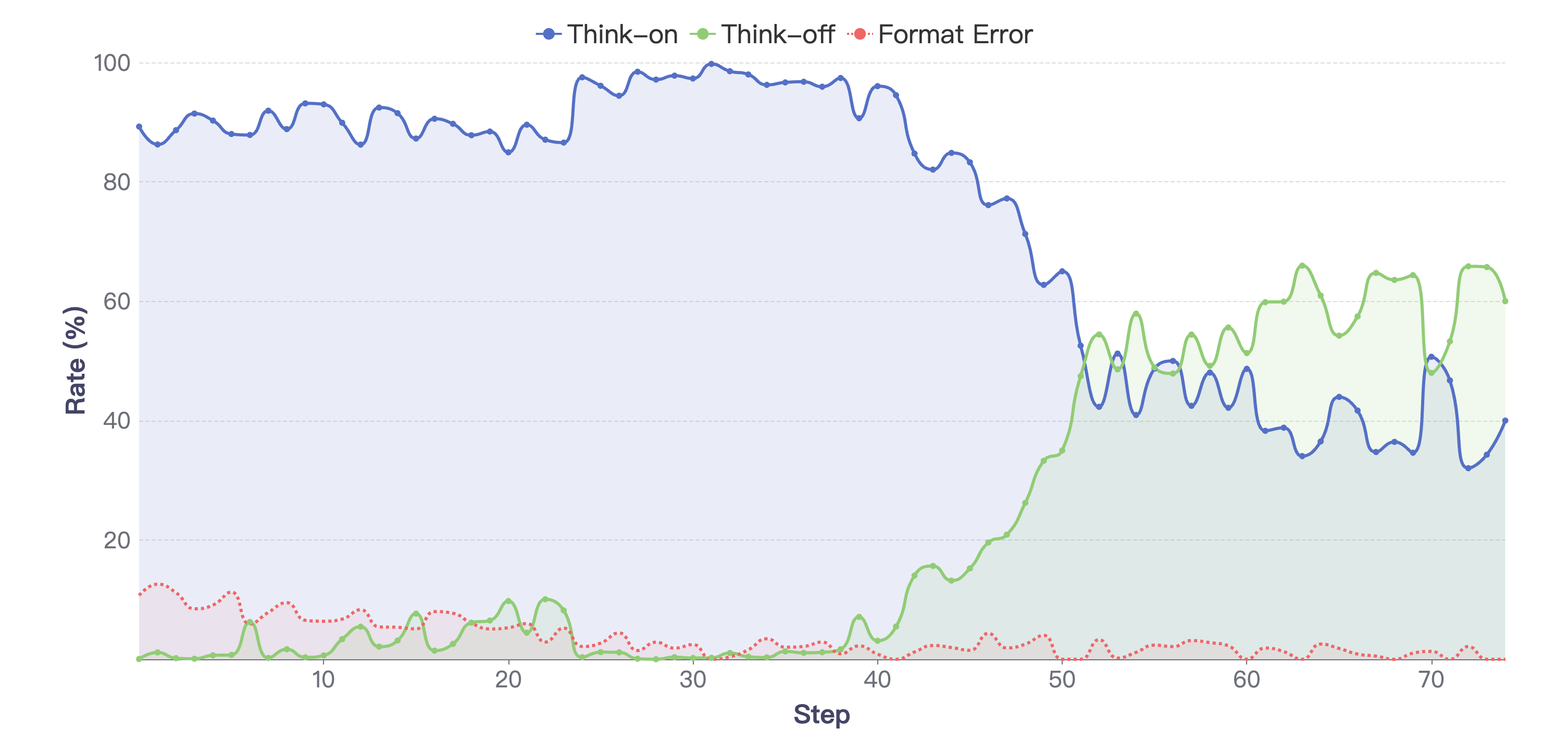}
    \caption{\textbf{Training dynamics of reasoning-mode selection under Step-SRPO.} We track the model's preference for \textless think\_on\textgreater{} (blue) versus \textless think\_off\textgreater{} (green) across training steps.}
    \label{fig:Step-SRPO-Rate}
\end{figure}

We logged how often the model chose <think\_on> versus <think\_off> at each step. This shift---visualized in Figure~\ref{fig:Step-SRPO-Rate}---confirms that Step-SRPO not only improves final accuracy but also refines the model's gating behavior, enabling it to skip heavy reasoning when unnecessary.

\begin{figure}[htbp]
    \centering
    \includegraphics[width=0.95\linewidth]{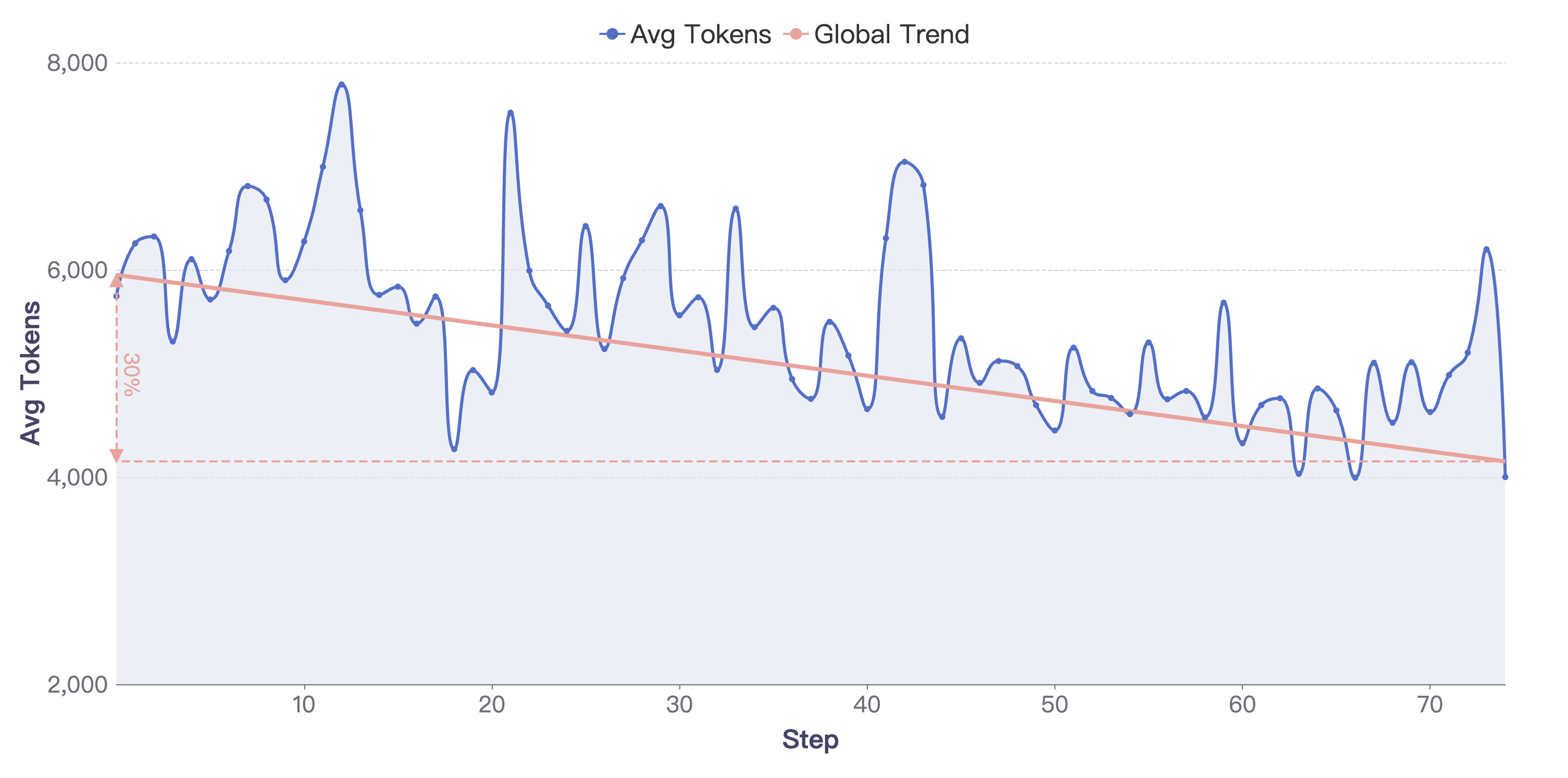}
    \caption{\textbf{Token usage trend during reinforcement learning in Step-SRPO.} During training, the average output token count steadily declines.}
    \label{fig:Step-SRPO-Token}
\end{figure}

During AutoThink RL training, the average output token count per prediction exhibits a clear downward trend (see Figure~\ref{fig:Step-SRPO-Token}). This reduction indicates that the model progressively learns to generate more concise responses, selectively invoking explicit reasoning only when necessary. Such behavior demonstrates the effectiveness of the Step-SRPO reward design in promoting efficient use of chain-of-thought and controlling response length across diverse queries.

\subsubsection{Think-On vs. Think-Off and Token Count Dynamics During Inference}

\begin{figure}[H]
    \centering
    \includegraphics[width=1\linewidth]{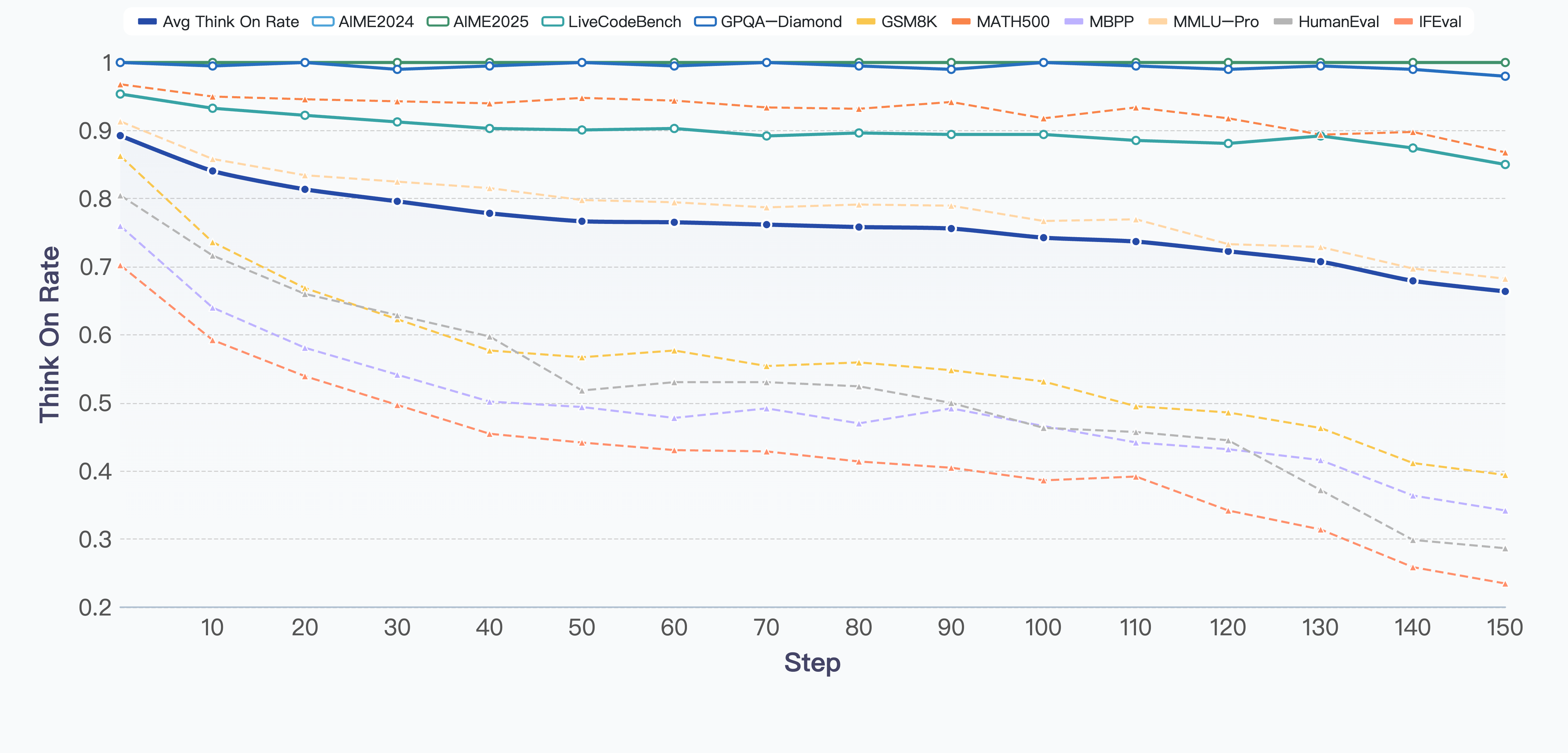}
    \caption{Think-on rate dynamics during inference across 10 benchmark tasks.}
    \label{fig:think-on-rate_inference}
\end{figure}

We systematically evaluated the evolution of reasoning control (think-on versus think-off activation) and token efficiency across multiple benchmark tasks. Figure \ref{fig:think-on-rate_inference} shows the proportion of think-on activations across different evaluation datasets during inference. Reasoning-intensive tasks, including AIME2024, AIME2025, GPQA-Diamond, and LiveCodeBench, consistently demonstrate high think-on activation rates (>80\%) throughout training. Conversely, tasks that require less explicit reasoning---such as GSM8K, MBPP, HumanEval, CEval, and IFEval---exhibit a clear downward trend in think-on activation as training progresses. Across all datasets, the average think-on rate decreased significantly from approximately 90\% to 66\%, representing a relative reduction of about 24\% and highlighting the model's enhanced capability for selectively invoking reasoning.

\begin{figure}[htbp]
    \centering
    \includegraphics[width=1\linewidth]{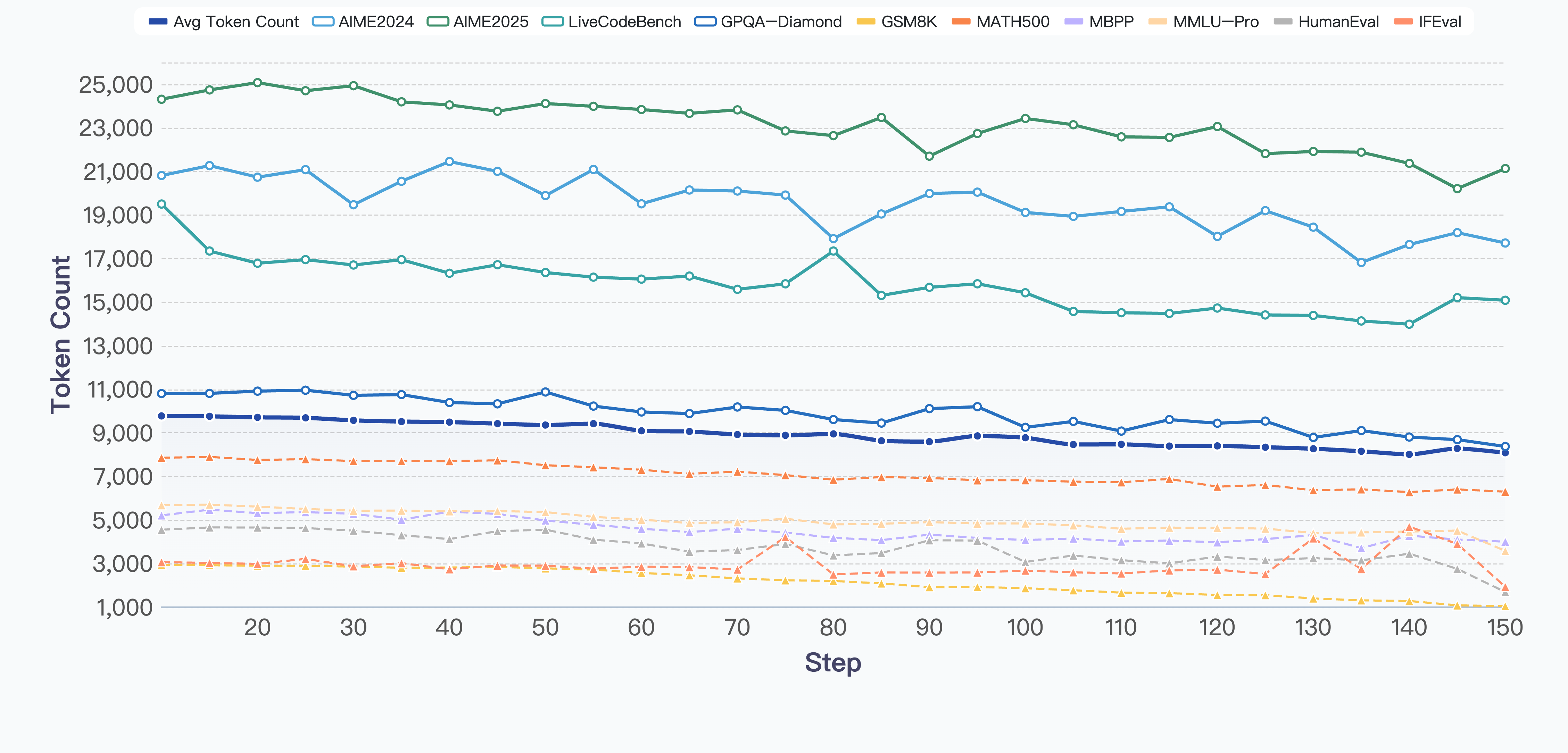}
    \caption{Token Count Dynamics during inference across 10 benchmark tasks.}
    \label{fig:tokenCountDynamics_inference}
\end{figure}

Consistent with reasoning mode trends, Figure \ref{fig:tokenCountDynamics_inference} illustrates the corresponding dynamics in average token counts per generated response during inference. Datasets with consistently high think-on activation maintain higher average token counts, indicative of sustained explicit reasoning. In contrast, datasets with reduced think-on activation show significant decreases in token usage over the course of training. Specifically, the average number of generated tokens across all datasets decreased from an initial 9,887 to 8,037 tokens , reflecting a relative reduction of approximately 19\% and underscoring improvements in inference efficiency achieved through adaptive reasoning control.

\section{Results}

To evaluate the effectiveness of KAT, we compared our model with several leading open-source state-of-the-art (SOTA) models, including LLaMA-4-Scout, LLaMA-4-Maverick, Qwen3-32B, Qwen3-235B-A22B, DeepSeek-V3, and DeepSeek-R1-0528 across a variety of benchmarks. As shown in Table~\ref{tab:main-comparison}, KAT consistently matches or outperforms these models in terms of accuracy across tasks in general reasoning, math and text reasoning, as well as agent and coding tasks.

\begin{figure}[H]
    \centering
    \includegraphics[width=0.95\linewidth]{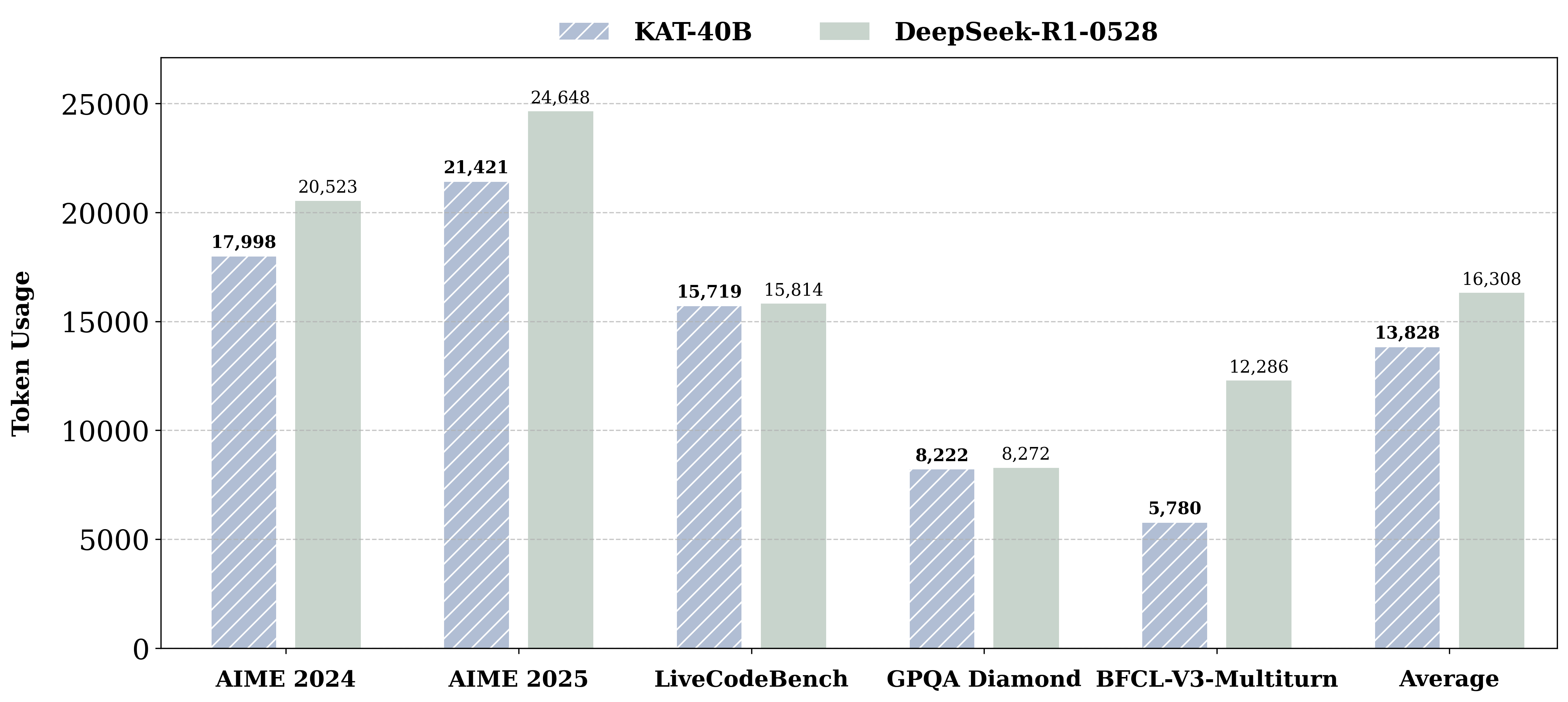}
    \caption{Comparison of Token Usage between KAT-V1-40B and DeepSeek-R1-0528.}
    \label{fig:token_usage_comparison}
\end{figure}

In addition to competitive accuracy, KAT demonstrates improvements in token efficiency. As shown in Figure~\ref{fig:token_usage_comparison}, across various benchmarks, token savings are consistently achieved.
These improvements are enabled by KAT's dynamic reasoning control, which selectively activates reasoning only when necessary, significantly reducing token usage in simpler queries. The model adapts its reasoning depth based on task complexity, resulting in fewer tokens consumed without sacrificing response quality.

In summary, Kwaipilot-AutoThink offers both high performance and enhanced efficiency, driven by a strong reasoning control strategy and an adaptive approach to token usage. This makes KAT a resource-efficient solution for real-world applications, as evidenced by its integration into Kwaipilot, Kuaishou's internal coding assistant.

\begin{table*}[htbp]
\centering
\renewcommand{\arraystretch}{0.85}
\caption{Comparison of KAT and leading SOTA models across diverse benchmark tasks. The numbers, enclosed in a \fbox{box}, \textbf{bolded}, and \underline{underlined}, respectively indicate the top three performance rankings of the models on the current benchmark.}
\label{tab:main-comparison}
\scalebox{1}{
\resizebox{\textwidth}{!}{
\begin{tabular}{
  >{\centering\arraybackslash}p{0.2\textwidth}
  *{8}{>{\centering\arraybackslash}p{0.1\textwidth}}
}
\toprule
& \makecell[c]{\textbf{~~~~~KAT~~~~~}\\\textbf{-V1-40B}}
& \makecell[c]{\textbf{~~~~~KAT~~~~~}\\\textbf{-V1-200B}}
& \makecell[c]{\textbf{DeepSeek}\\\textbf{-R1-0528}}
& \makecell[c]{\textbf{DeepSeek}\\\textbf{-V3}}
& \makecell[c]{\textbf{Qwen3}\\ \textbf{-235B-A22B}}
& \makecell[c]{\textbf{~~~Qwen3~~~}\\\textbf{-32B}}
& \makecell[c]{\textbf{LLaMA-4}\\\textbf{-Maverick}}
& \makecell[c]{\textbf{LLaMA-4}\\\textbf{-Scout}} \\

\midrule
Architecture         & Dense & MoE & MoE & MoE & MoE & Dense & MoE & MoE \\
\# Total Params      & 40B & 200B & 671B & 671B & 235B & 32B & 400B & 109B \\
\# Act Params  & 40B & 40B & 37B & 37B & 22B & 32B & 17B & 17B \\
\midrule
\multicolumn{9}{c}{\textit{General Tasks}} \\
\cmidrule(lr){1-9}
MMLU-Pro\cite{wang2024mmlu-pro}     & 77.8 & \fbox{82.3} & \textbf{81.9} & \underline{81.2} & 80.8 & 76.2 & 80.5 & 74.3 \\
DROP\cite{Dua2019DROP}          & 91.2 & \underline{91.6} & \fbox{92.2} & 90.5 & \textbf{91.9} & 91.1 & 90.7 & 89.2 \\
WildBench\cite{lin2024wildbench}     & 8.73 & \textbf{8.77} & \fbox{8.80} & \underline{8.74} & 8.72 & 8.56 & 8.29 & 8.21 \\
GPQA-D\cite{rein2024gpqa-Diamond}  & \underline{75.1} & \textbf{78.2} & \fbox{81.0} & 68.4 & 71.1 & 68.4 & 69.8 & 57.2 \\
\midrule
\multicolumn{9}{c}{\textit{Math \& Text Reasoning}} \\
\cmidrule(lr){1-9}
MATH500\cite{huggingface-math500}       & \underline{97.4} & \textbf{97.6} & 97.3 & 94.0 & \fbox{98.0} & 97.2 & 90.6 & 82.6 \\
AIME2024\cite{huggingface-aime2024}      & \textbf{93.3} & \fbox{96.3} & \underline{91.4} & 59.4 & 85.7 & 81.4 & 38.5 & 28.6 \\
AIME2025\cite{huggingface-aime2025}      & \textbf{88.1} & \fbox{90.4} & \underline{87.5} & 44.2 & 81.5 & 72.9 & 15.9 & 10.0 \\
AutoLogi\cite{zhu2025autologi}      & 82.1 & \underline{84.9} & 84.3 & 76.1 & \fbox{89.0} & \textbf{87.3} & 75.2 & 56.8 \\
\midrule
\multicolumn{9}{c}{\textit{Agent \& Coding Tasks}} \\
\cmidrule(lr){1-9}
HumanEval\cite{chen2021openai-humaneval}     & \underline{95.1} & \fbox{98.2} & \textbf{97.0} & 93.3 & 30.5 & 90.2 & 88.4 & 83.5 \\
MBPP\cite{austin2021mbpp}          & \underline{85.2} & \textbf{93.2} & \fbox{96.2} & 83.6 & 65.4 & 74.6 & 81.0 & 72.0 \\
LiveCodeBench\cite{jain2024livecodebench} & \underline{73.1} & \fbox{77.1} & \textbf{73.3} & 44.7 & 59.1 & 61.5 & 34.1 & 30.2 \\
LCB-Pro-Med\cite{liveCodeBench-Pro} & \fbox{12.7} & N/A & \textbf{7.0} & 0.0 & N/A & N/A & 0.0 & N/A \\
BFCL-V3-Mul\cite{patil2025bfcl-v3} & 37.4 & \textbf{41.6} & 38.9 & 29.9 & \underline{40.1} & \fbox{43.1} & 16.4 & 1.9 \\
\bottomrule
\end{tabular}
}
}
\end{table*}

\section{Work with KwaiPilot: Real-World Applications}

Beyond strong benchmark performance, KAT-V1-40B has also demonstrated its practical value through deployment in Kwaipilot, Kuaishou's internal coding assistant. To validate the real-world applicability of our thinking control mechanism and training pipeline, we conducted a series of controlled product-level evaluations. 

\subsection{Avoiding Overthinking in Simple Queries}

\begin{figure}[H]
    \centering
    \includegraphics[width=1\linewidth]{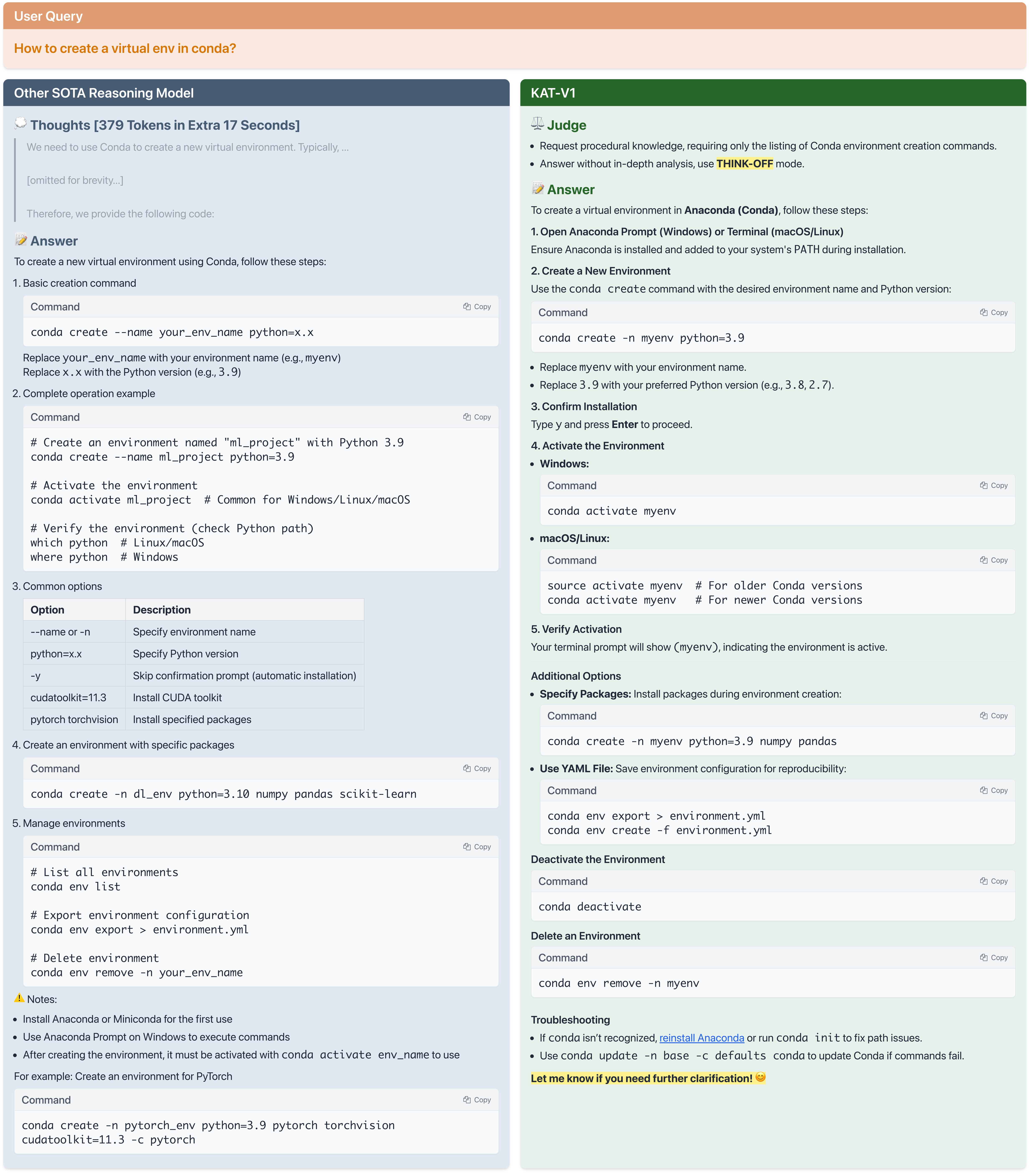}
    \caption{Comparison between KAT-V1-40B and a state-of-the-art reasoning model on a simple procedural query.}
    \label{fig:avoid-thinking}
\end{figure}

As shown in Figure~\ref{fig:avoid-thinking}, when handling a query that does not require explicit reasoning, the state-of-the-art reasoning model still engages in unnecessary step-by-step analysis, generating a verbose response of nearly 400 tokens and incurring an additional 17 seconds of latency. In contrast, KAT-V1-40B correctly recognizes the simplicity of the task, promptly activates the think-off mode, and produces a high-quality response.

This example underscores KAT-V1-40B's ability to avoid overthinking in low-complexity scenarios, significantly improving response efficiency while reducing token consumption. Moreover, thanks to the multi-agent generated reasoning data introduced during the knowledge enhancement stage, KAT-V1-40B exhibits strong interaction capabilities and a fine-grained understanding of when reasoning is contextually appropriate, further reinforcing its practical value in real-world deployments.

\subsection{Reasoning Mode for Complex Query}

\begin{figure}[H]
    \centering
    \includegraphics[width=1\linewidth]{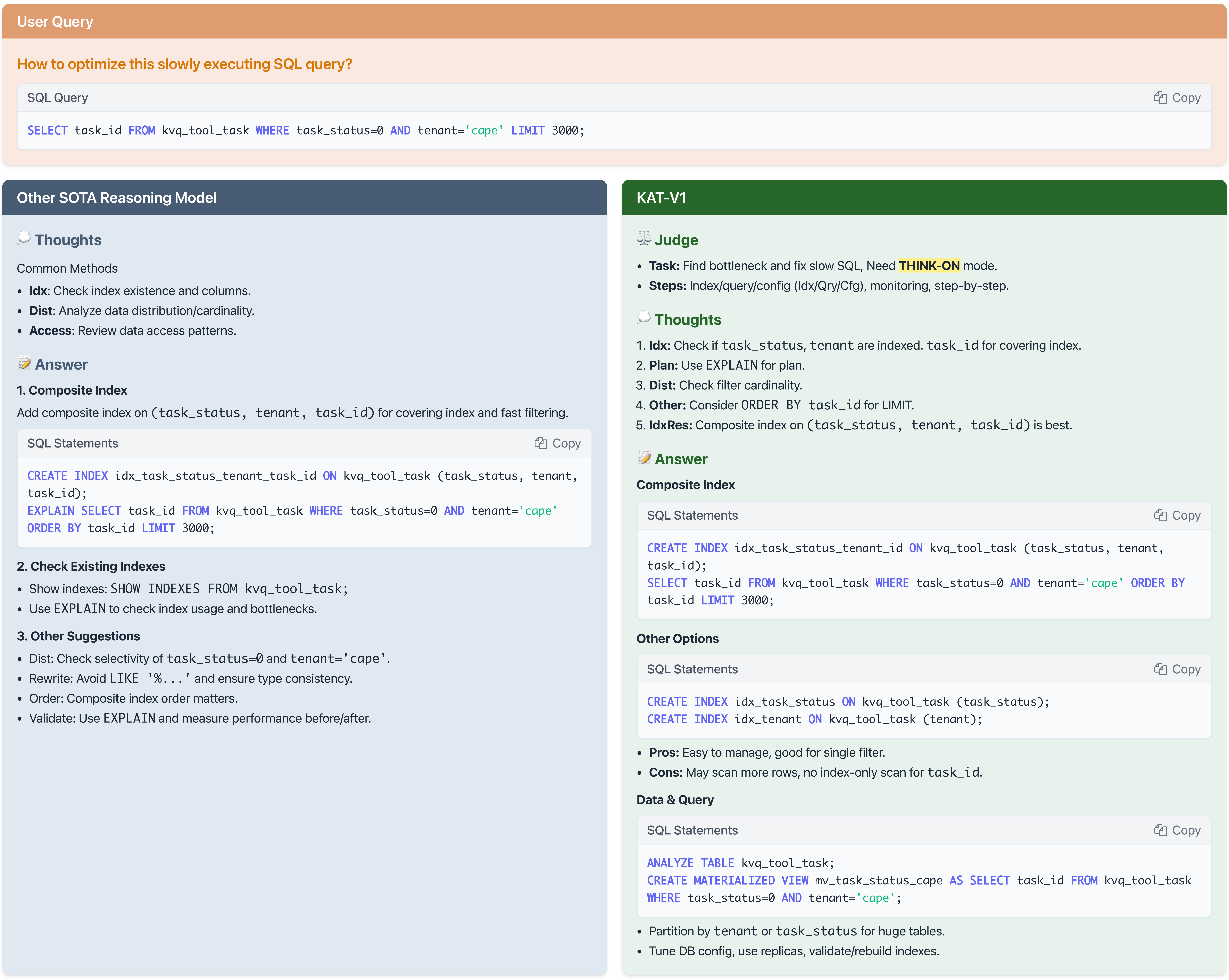}
    \caption{Comparison between KAT-V1-40B and a state-of-the-art reasoning model on a complex SQL performance tuning task.}
    \label{fig:reasoning-sql}
\end{figure}

When faced with a more complex scenario involving SQL performance tuning (Figure~\ref{fig:reasoning-sql}), KAT-V1-40B automatically engages its reasoning mode (think-on). It delivers a structured, multi-step analysis within just 15 seconds of thinking, while another reasoning model we host took 53 seconds. The response is rated on par with the state-of-the-art reasoning model in correctness, but outperforms it in depth, architectural insight, and scalability suggestions. This demonstrates KAT-V1-40B's capacity to adjust its reasoning depth dynamically and efficiently in real-world engineering tasks.

\subsection{End-to-End Mode Switching with Multi-Turn Capability}

\begin{figure}[H]
    \centering
    \includegraphics[width=1\linewidth]{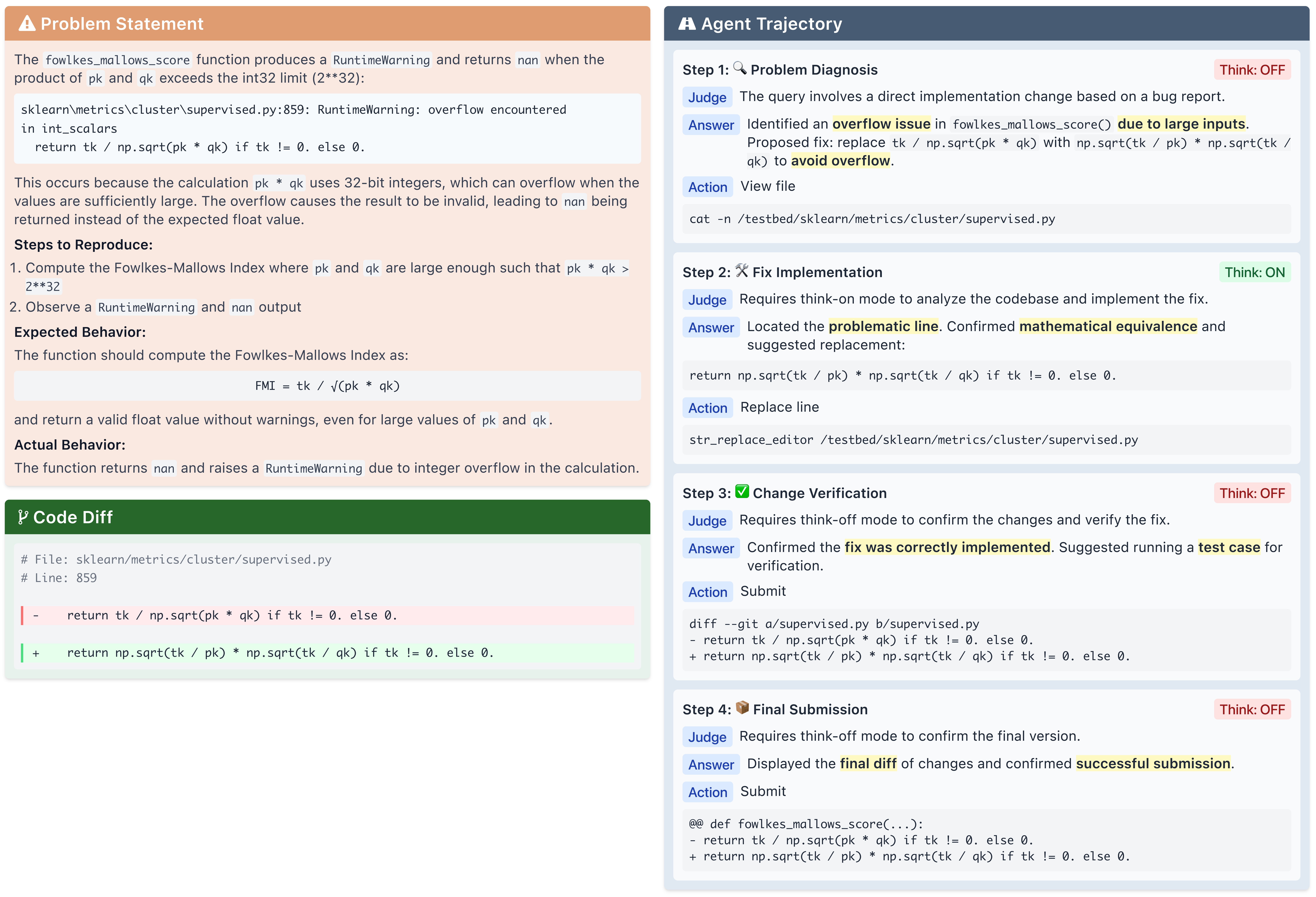}
    \caption{Example of KAT-V1-40B executing a full multi-turn agent trajectory while dynamically switching between reasoning modes.}
    \label{fig:example-multi-turn}
\end{figure}

As shown in Figure~\ref{fig:example-multi-turn}, KAT-V1-40B demonstrates its end-to-end controllability over reasoning mode, accurately alternating between think-off and think-on across different steps in a multi-turn agent trajectory. It passively disables reasoning during file inspection (think-off) and actively enables deep reasoning and tool-based exploration (think-on) when diagnosis or code generation is required. The agent exhibits tool-using proficiency, context retention, and accurate task decomposition across multiple interaction rounds, validating KAT-V1-40B's ability to autonomously manage complex task flows in production settings.

\subsection{User-Guided Mode Switching}

In addition to autonomous reasoning control, KAT-V1-40B also supports user-guided control over reasoning mode, allowing users to flexibly override the model's default behavior. By embedding simple intent directives, such as explicit reasoning instructions or non-reasoning preferences, into the input query, users can directly influence whether the model engages in think-on or think-off mode.

\begin{figure}[]
    \centering
    \includegraphics[width=1\linewidth]{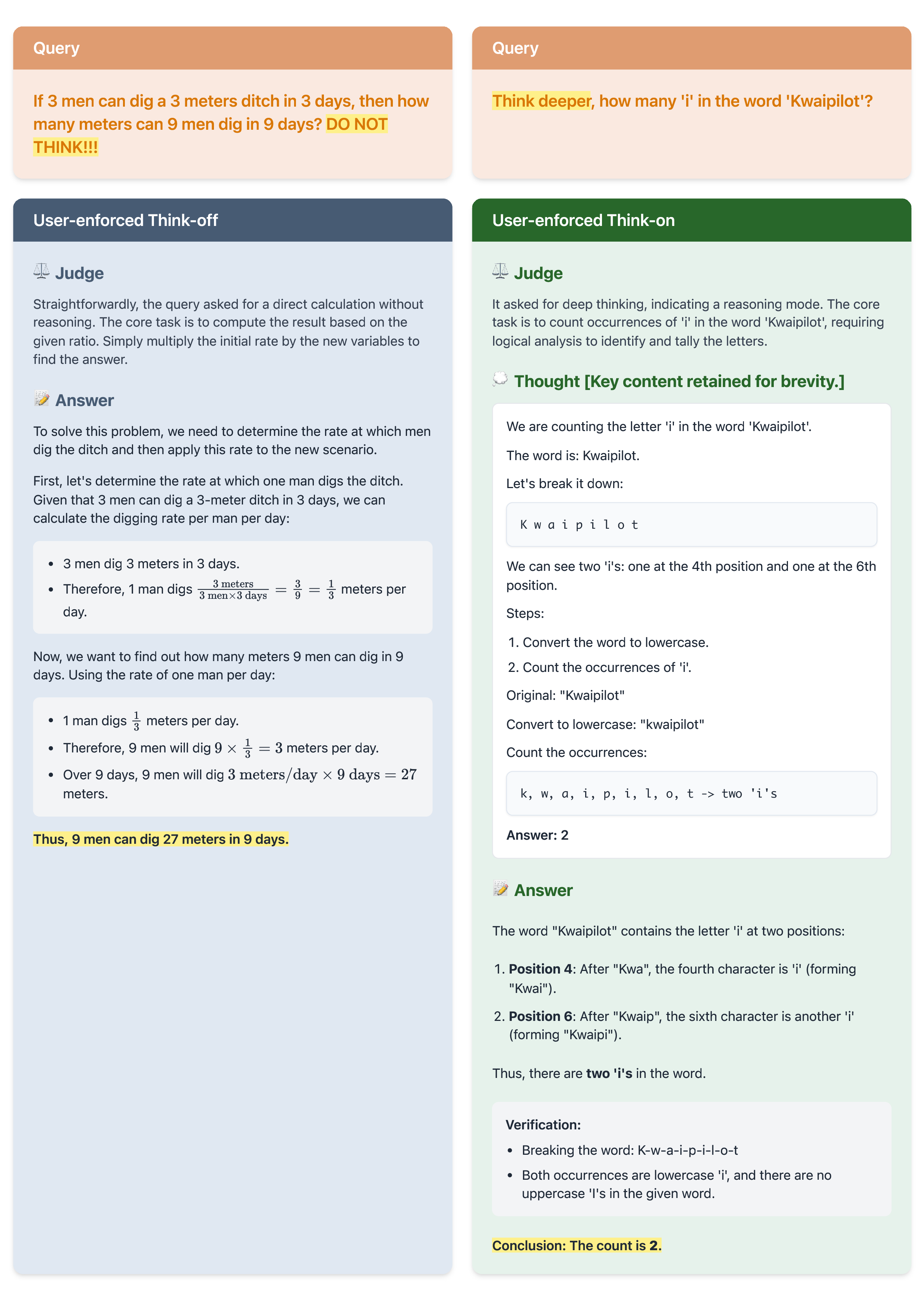}
    \caption{Examples of user-guided mode switching in KAT-V1-40B.}
    \label{fig:user-guided_mode_switching}
\end{figure}

As illustrated in Figure~\ref{fig:user-guided_mode_switching}, on the left, the user enforces the non-reasoning mode with the directive "DO NOT THINK," leading KAT-V1-40B to skip reasoning steps and directly compute the answer. On the right, the instruction "Think deeper" prompts the model to engage in step-by-step reasoning. These examples demonstrate KAT-V1-40B's ability to interpret and follow user-specified intent for reasoning depth, enabling flexible and controllable interaction.

This user-in-the-loop controllability further enhances KAT-V1-40B's practicality in real-world systems, enabling flexible trade-offs between efficiency and interpretability based on task requirements or user preferences. It also provides a natural interface for integrating KAT-V1-40B into interactive workflows, where explicit control over reasoning depth is desirable.

\subsection{Insights from Real-World Use Cases}

These case studies collectively demonstrate that Kwai-AutoThink is not only competitive with state-of-the-art models on standard benchmarks but also excels in real-world production environments. Through comprehensive evaluations across diverse coding tasks---including but not limited to SQL optimization and agent-style multi-turn interactions---KAT-V1-40B has shown the ability to dynamically select appropriate reasoning modes based on task complexity, achieving strong performance in both response quality and efficiency.

In addition to its autonomous reasoning control, KAT-V1-40B supports user-guided thinking mode switching via prompt-level intent specification, giving users explicit control over reasoning depth when needed. This dual-mode flexibility, adaptive when desired and controllable when required, further enhances its utility in practical deployments.
In general, these findings highlight the deployability, efficiency, and controllability of our thinking control framework, validating KAT-V1-40B not just as a research prototype, but as a solid foundation for building scalable, high-performance, and production-ready reasoning systems.

\section{Conclusion}

In this work, we present Kwaipilot-AutoThink, a 40B open-source large language model designed to address the overthinking problem in reasoning-intensive tasks. We propose a novel automatic thinking training framework that enables dynamic switching between reasoning and non-reasoning modes, guided by task complexity. Our approach combines a diverse dual-regime data synthesis pipeline, MTP-enhanced knowledge distillation, and Step-SRPO, a staged reinforcement learning algorithm with intermediate supervision. Together, these components allow the model to learn fine-grained reasoning behavior with minimal pretraining cost.

Through extensive experiments across multiple benchmarks, KAT achieves state-of-the-art performance while significantly reducing token usage---matching or exceeding models with several times more parameters. Furthermore, the model has been successfully deployed in real-world production, demonstrating strong practical value in Kuaishou's internal coding assistant, Kwaipilot.

Our findings highlight the importance of adaptive reasoning control, efficient knowledge transfer, and structured supervision for building scalable, high-performance, and deployable LLMs. We believe KAT offers a promising direction toward closing the gap between academic progress and real-world usability. 

\section{Future Works}

Looking ahead, we will release a companion paper detailing our full AutoThink training framework, including cold-start initialization and reinforcement learning strategies. We will also open-source the associated training data, RL codebase, and a suite of models ranging from 1.5B, 7B to 13B parameters. Additionally, we plan to open-source our 200B-parameter Mixture-of-Experts (MoE) model upon completion of training, offering the community a high-capacity, sparsely activated LLM built on the AutoThink foundation.

Beyond the current scope, we aim to extend AutoThink to multi-modal and interactive agent settings, further enhancing the controllability, generality, and applicability of reasoning-capable language models in complex real-world environments.

\newpage
\bibliography{main}
\end{document}